\pdfoutput=1
\documentclass[11pt]{article}

 \usepackage{emnlp2021}

\usepackage{times}
\usepackage{latexsym}

\usepackage[T1]{fontenc}
\usepackage[utf8]{inputenc}

\usepackage{microtype}

\usepackage{amsmath,amssymb,amsfonts}
\usepackage{eurosym}
\usepackage{algorithmic}
\usepackage{graphicx}
\usepackage{textcomp}
\usepackage{xcolor}
\usepackage{multirow}
\usepackage{tabularx,ragged2e}
\usepackage{tabulary}
\usepackage{adjustbox}
\usepackage{arydshln}
\usepackage{cleveref}
\usepackage[flushleft]{threeparttable}
\usepackage{xurl}
\usepackage{hhline}
\usepackage[group-separator={,}]{siunitx}
\usepackage{multicol}
\usepackage{booktabs}
\usepackage{footmisc}
\usepackage{graphicx}
\usepackage{arydshln}
\usepackage{calc}
\usepackage{svg}
\usepackage{float}
\usepackage{amssymb}
\usepackage{comment}
\PassOptionsToPackage{hyphens}{url}\usepackage{hyperref}
\title{Neural Media Bias Detection Using Distant Supervision With BABE - Bias Annotations By Experts}

\author{Timo Spinde\textsuperscript{1}, Manuel Plank\textsuperscript{2}, Jan-David Krieger\textsuperscript{2}, Terry Ruas\textsuperscript{1}, Bela Gipp\textsuperscript{1}, Akiko Aizawa\textsuperscript{3} \\
  \textsuperscript{1}University of Wuppertal, \textsuperscript{2} University of Konstanz, \textsuperscript{3}
  NII Tokyo\\
  \texttt{\{firstname.lastname\}@\{uni-wuppertal.de,@uni-konstanz.de\}} \\ \texttt{aizawa@nii.ac.jp}
}

\begin{document}
\maketitle
\newpage
%optional re-enable header on the second page (do not execute this command on the first page)
%\pagestyle{standardpagestyle}
%%%%%%%%%%%%%%%%%%%%%%%%%%%%%%%%%%%%%%%%%%%%%%%%%%%%%%%%%%
%%% ABSTRACT
%%%%%%%%%%%%%%%%%%%%%%%%%%%%%%%%%%%%%%%%%%%%%%%%%%%%%%%%%%

\begin{abstract}
 Media coverage has a substantial effect on the public perception of events. Nevertheless, media outlets are often biased. One way to bias news articles is by altering the word choice. The automatic identification of bias by word choice is challenging, primarily due to the lack of a gold standard data set and high context dependencies. This paper presents BABE, a robust and diverse data set created by trained experts, for media bias research. We also analyze why expert labeling is essential within this domain. Our data set offers better annotation quality and higher inter-annotator agreement than existing work. It consists of 3,700 sentences balanced among topics and outlets, containing media bias labels on the word and sentence level. Based on our data, we also introduce a way to detect bias-inducing sentences in news articles automatically. Our best performing BERT-based model is pre-trained on a larger corpus consisting of distant labels. Fine-tuning and evaluating the model on our proposed supervised data set, we achieve a macro $F_{1}$-score of 0.804, outperforming existing methods.
\end{abstract}

%%%%%%%%%%%%%%%%%%%%%%%%%%%%%%%%%%%%%%%%%%%%%%%%%%%%%%%%%%
%%% INTRODUCTION
%%%%%%%%%%%%%%%%%%%%%%%%%%%%%%%%%%%%%%%%%%%%%%%%%%%%%%%%%%

\section{Introduction}\label{Intro}
Online news articles have, over time, started to replace traditional print and radio media as a primary source of information \cite{dallmann2015a}. A varying word choice may have a major effect on the public and individual perception of societal issues, especially since regular news consumers are mostly not fully aware of the degree and scope of bias \cite{spinde2020b}. As shown in existing research~\cite{park2009newscube, baumer2015a}, detecting and highlighting media bias might be relevant for media analysis and to mitigate the effects of biased reports on readers. Also, the detection of media bias can assist journalists and publishers in their work \cite{Spinde2021}. 
To date, only a few research projects focus on the detection and aggregation of bias \cite{lim2020annotating, Spinde2020INRA}. Even though bias embodies a complex structure, contributions \cite{hube2019neural, chen_analyzing_2020} often neglect annotator background and use crowdsourcing to collect annotations. Therefore, existing data sets exhibit low annotator agreement and inferior quality.

Our study holds both theoretical and practical significance. We propose BABE (\textbf{B}ias \textbf{A}nnotations \textbf{B}y \textbf{E}xperts), a data set of media bias annotations, which is built on top of the MBIC data set \cite{spinde2021mbic}. MBIC offers a balanced content selection, annotations on a word and sentence level, and is with 1,700 annotated sentences one of the largest data sets available in the domain. BABE improves MBIC, and other data sets, in two aspects. First, annotations are performed by trained experts and in a larger number. Second, the corpus size is expanded considerably with additional 2,000 sentences.  The resulting labels are of higher quality and capture media bias better than labels gathered via crowdsourcing. In sum, BABE consists of 3,700 sentences with gold standard expert annotations on the word and sentence level.\footnote{We also provide another 1,000 yet unlabeled sentences for future work. We have not labeled them to date due to resource restrictions.}

To analyze the ideal trade-off between the number of sentences, annotations, and human annotation cost, we divide our gold standard into 1,700 and 2,000 sentences, which are annotated by eight and five experts, respectively.\footnote{With the 1,700 stemming from MBIC \cite{spinde2021mbic}.} Lastly, we train and present a neural BERT-based classifier that outperforms existing approaches such as the one by \citet{Spinde2021}. Even though neural network architectures have been applied to the media bias domain \cite{hube2019neural,chen_analyzing_2020}, their data sets created using crowdsourcing do not exhibit similar quality as our expert data set. In addition, we include five state-of-the-art neural models in our comparison and extend two of them in a distant supervision approach \cite{tang2014,deriu2017}. Leveraging large amounts of distantly labeled data, we formulate a pre-training task helping the model to learn bias-specific embeddings by considering bias information when optimizing its loss function. For the classification presented in this paper, we focus on sentence level bias detection, which is the current standard in related work (Section \ref{sec:relatedwork})\footnote{Our data set is in English language, which is also currently most common in the domain \cite{Spinde2021e}.}. We address future work on word level bias in Section \ref{sec:discussion}. We publish all our code and resources on \url{https://github.com/Media-Bias-Analysis-Group/Neural-Media-Bias-Detection-Using-Distant-Supervision-With-BABE}.

%%%%%%%%%%%%%%%%%%%%%%%%%%%%%%%%%%%%%%%%%%%%%%%%%%%%%%%%%%
%%% RELATED WORK
%%%%%%%%%%%%%%%%%%%%%%%%%%%%%%%%%%%%%%%%%%%%%%%%%%%%%%%%%%

\section{Related Work}\label{sec:relatedwork}
Media bias can be defined as slanted news coverage or internal news article bias \cite{recasens2013a}. While there are multiple forms of bias, e.g., bias by personal perception or by the omission of information \cite{PUGLISI2015647}, our focus is on bias caused by word choice, in which different words refer to the same concept. For a detailed explanation of the types of media bias, we refer to \citet{Spinde2021}. In the following, we summarize the existing literature on bias data sets and media bias classification. 

\subsection{Media Bias Data Sets}\label{sec:data sets}
\citet{lim2018b} present 1,235 sentences labeled for word and sentence level bias by crowdsource workers. All the sentences in their data set focus on one event. Another data set focusing on just one event is presented by \citet{10.1145/3340531.3412876}. It consists of 2,057 sentences from 90 news articles, annotated with bias labels on article and sentence levels, and contains labels such as overall bias, hidden assumption, and framing. The annotators agree with a Krippendorff's $\alpha$ = -0.05. \citet{lim2020annotating} also provide a second data set with 966 sentences labeled on the sentence level. However, their reported interrater-agreement (IRR) of Fleiss' Kappa on different topics averages at zero. 

\citet{baumer2015a} classify framing in political news. Using crowdsourcing, they label 74 news articles from eight US news outlets, collected from politics-specific RSS feeds on two separate days. \citet{chen_analyzing_2020} create a data set of 6,964 articles containing political bias, unfairness, and non-objectivity labels at the article level. Altogether, they present 11 different topics such as “presidential election”, “politics”, and “white house”.

\citet{fan2019a} present 300 news articles containing annotations for lexical and informational bias made by two experts. They define lexical bias as bias stemming from specific word choice, and informational bias as sentences conveying information tangential or speculative to sway readers’ opinions towards entities \cite{fan2019a}. Their data set, BASIL, allows for analysis at the token level and relative to the target, but only 448 sentences are available for lexical bias.

Under the name MBIC, \citet{spinde2021mbic} extract 1,700 sentences from 1,000 news articles. Crowdsource workers then label bias and opinion on a word and sentence level using a survey platform that also surveyed the annotators' backgrounds. MBIC covers 14 different topics and yields a Fleiss' Kappa score of 0.21. 

Even though the referenced data sets contribute valuable resources to the media bias investigation, they still have significant drawbacks, such as (1) a small number of topics \cite{lim2018b, lim2020annotating}, (2) no annotations on the word level \cite{lim2018b}, (3) low inter-annotator agreement \cite{spinde2021mbic, lim2020annotating, baumer2015a, lim2018b}, and (4) no background check for its participants (except \cite{spinde2021mbic}). Also, some related papers focus on framing rather than on bias \cite{baumer2015a, fan2019a}, and results are only partially transferable. Our work aims to address these weaknesses by gathering sentence level annotations about bias by word choice over a balanced and broad range of topics. The annotations are made by trained expert annotators with a higher capability of identifying bias than crowdsource workers. 

\subsection{Media Bias Classification Systems}
Several studies tackle the automated detection of media bias \cite{hube2018detecting, spinde2020a, chen_analyzing_2020}. Most of them use manually created features to detect bias \cite{hube2018detecting}, and are based on traditional machine learning models \cite{Spinde2021}.  

\citet{recasens2013a} identify sentence level bias in Wikipedia using supervised classification. They use a bias lexicon and a set of various linguistic features (e.g., assertive verbs, sentiment) with a logistic regression classifier, identifying bias-inducing words in a sentence. They also report that crowdsource workers struggle to identify bias words that their classifier is able to detect. 

\citet{Spinde2021} create a media bias data set (i.e., MBIC) and develop a feature-based tool to detect bias-inducing words. The authors identify and evaluate a wide range of linguistic, lexical, and syntactic features serving as potential bias indicators. Their final classifier returns an $F_{1}$-score of 0.43 and 0.79 AUC. Spinde et al. point out the explanatory power of various feature-based approaches and the performance of their own model on the MBIC data set. Yet, their results indicate that Deep Learning models are promising alternatives for future work. 
%\citet{Spinde2021} claim the subjective complexity of bias affects its detection by crowdsource workers without proper experience and training.

\citet{hube2018detecting} propose a semi-automated approach to extract domain-related bias based on word embeddings properties. 
The authors combine bias words and linguistic features  (e.g., report verbs, assertive verbs) in a random forest classifier to detect sentence level bias in Wikipedia.
They achieve an $F_{1}$-score of 0.69 on a newly created ground truth based on Conservapedia.\footnote{\url{https://conservapedia.com/Main\_Page}, accessed on 2021-04-10.} In their following work, \citet{hube2019neural} propose a neural statement-level bias detection approach based on Wikipedia data. Using recurrent neural networks (RNNs) and different attention mechanisms, the authors achieve an $F_{1}$-score of 0.77, indicating a possible advantage of neural classifiers in the domain.
\citet{chen_analyzing_2020} train a RNN to classify article-level bias. They also conduct a reverse feature analysis and find that, at the word level, political bias correlates with categories such as negative emotion, anger, and affect.

To summarize, most approaches use manually created features, leading to lower performance and poor representation. The few existing contributions on neural models are based on naive data sets  (cf. Section \ref{sec:data sets}). Therefore, we decided to develop a neural classifier trained on BABE. Our system incorporates state-of-the-art models and improves their pre-training step through distant supervision \cite{tang2014,deriu2017}, allowing the model to learn bias-specific embeddings, thus improving its representation. Almost all models focus on sentence level bias, describing it as the lowest meaningful level that can be aggregated to higher levels, like the document level. Therefore, we follow the standard practice and construct a sentence level classifier.

%%%%%%%%%%%%%%%%%%%%%%%%%%%%%%%%%%%%%%%%%%%%%%%%%%%%%%%%%%
%%% data set CREATION
%%%%%%%%%%%%%%%%%%%%%%%%%%%%%%%%%%%%%%%%%%%%%%%%%%%%%%%%%%

\section{Data Set Creation}\label{sec:data}

Since media bias by word choice rarely depends on context outside the sentences \cite{fan2019a}, we focused on gathering sentences only. To tackle the weaknesses of existing bias data sets, we created a robust and diverse corpus containing \textbf{B}ias \textbf{A}nnotations \textbf{B}y \textbf{E}xperts (BABE).

\subsection{Data Collection}
\begin{figure}[H]
    \centering
    \small 
    \def\svgscale{0.905}
    \input{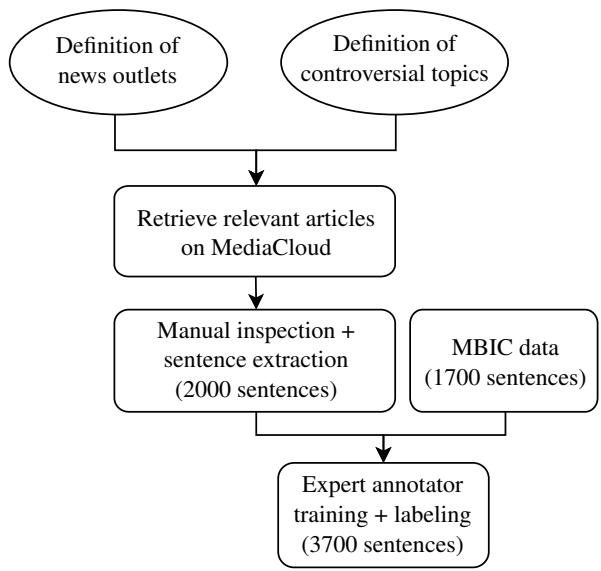}
    \caption{\label{workflow}Data collection and annotation pipeline}
\end{figure}

 The general data collection and annotation pipeline is outlined in Figure \ref{workflow}. Similar to the filtering strategy proposed by \citet{Spinde2021}, the sentences should contain more biased than neutral sentences. BABE contains 3,700 sentences, 1,700 from MBIC \cite{spinde2021mbic} and additional 2,000. Like \citet{spinde2021mbic}, we extracted our sentences from news articles covering 12 predefined controversial topics.\footnote{The list of topics is provided at the repository mentioned in Section \ref{Intro}.} The articles were published on 14 US news platforms from January 2017 until June 2020. We focused on the US media since their political scenario became increasingly polarizing over the last years \cite{atkins2016skewed}.
 
We selected appropriate left-wing, center, and right-wing news outlets based on the media bias chart provided by Allsides.\footnote{\url{https://www.allsides.com/media-bias/media-bias-chart}, accessed on 2021-04-13.} The sentence collection was performed on the open-source media analysis platform Media Cloud.\footnote{\url{https://mediacloud.org/}, accessed on 2021-04-13.} The collection process was as follows. We defined keywords describing every topic in one word or a short phrase, specified the news outlets, their time frame, and retrieved all available links for the relevant articles.\footnote{The keywords can be found at the repository mentioned in Section \ref{Intro}.} Then, we extracted sentences by manually inspecting the provided list of articles. The sentence selection was based on our media bias annotation guidelines comprising diverse examples of biased and neutral text instances (see Section \ref{sec:data_annotation}).

\subsection{Data Annotation} \label{sec:data_annotation}
As laid out in Section \ref{sec:relatedwork}, high-quality annotations are often obtained if the participants are properly instructed and have sufficient training \cite{fan2019a,Spinde2021}.  We compare our expert annotations with the crowdsourced labels provided by \citet{spinde2021mbic} to further analyze quality differences between the two groups. Our results show that expert annotators render more qualitative bias labels than MBIC's crowdsourcers.

We define as an expert a person with at least six months of experience in the media bias domain and underwent sufficient training to (1) reliably identify biased wording, (2) distinguish between bias and plain polarizing language, and  (3) take on a politically neutral viewpoint when annotating.\footnote{Note: We cannot guarantee that a media bias expert is fully neutral, but we assume that an expert is able to leave political viewpoints aside to a substantial extent.} To build up such experience, we developed detailed instruction guidelines that are presented before the annotation task.\footnote{%Inspired by \cite{spinde2021b}%, 
Available on the repository mentioned in Section \ref{Intro}.} The instructions are substantially more comprehensive than instructions in a crowdsourcing setting.
Considering that the annotation of bias on a fine-grained linguistic level is a complex task, and cognitive and language abilities likely have an impact on text perception \cite{kause2019framing}, we hired only master students from programs completely held in English, who were among the top 20\% with respect to their grade. 
Based on an iterative feedback loop between all annotators and us, we refined the guidelines multiple times with richer and clearer details. 
We discussed and evaluated existing annotations weekly as a group during the first three weeks of each annotator's work.
We also always asked each annotator to hand in annotations before the discussion sessions, so they could not influence each other. The annotators had to provide basic reasoning about their annotation decisions during our discussions. We maintained the labels only if the annotators were able to elaborate their annotations. Annotations of one annotator were discarded based on this method. 

Apart from evaluation and instructions, each annotator rated at least 1,700 sentences to improve experience over time.\footnote{The same sentences as in MBIC.} On average, per hour, they were paid 15,00\texteuro  ~and labeled 40 sentences, costing approximately 10,000\texteuro . The sum of money required to obtain a sufficient number of reasonable bias labels can be restrictive for media bias research. Therefore, BABE represents a major contribution that alleviates the lack of high-quality annotations in the domain. The annotators were instructed to label carefully and not as fast as possible, even though this resulted in a higher overall cost.

The general instructions for the annotation task were identical to the approach by \citet{spinde2021mbic}. First, raters were asked to mark words or phrases inducing bias. Then, we asked them to indicate whether the whole sentence was \textit{biased} or \textit{non-biased}. Lastly, the annotators labeled the sentence as \textit{opinionated}, \textit{factual}, \textit{or mixed}.

As our resources were limited and the ideal trade-off between the number of sentences and annotators per sentence is not yet determined, we organized BABE into subgroups (SG), as described below: 
\begin{itemize}
    \item \textbf{SG1}. 1,700 sentences annotated by eight expert raters each.
    \item \textbf{SG2}. 3,700 sentences annotated by five expert raters each.
\end{itemize}

 For SG1, we hired eight raters to annotate the 1,700 sentences (same as MBIC) on word and sentence levels \citep{spinde2021mbic}.\footnote{In the original MBIC data set, each sentence was evaluated by ten crowdsource workers \cite{spinde2021mbic}.} Thereby, we obtained an expert-labeled ground truth comparable to MBIC's crowdsourcing results. For SG2, five of the previous eight annotators also labeled the 2,000 additionally collected sentences. We explored the ideal number of annotators by sampling. 5 annotators is a compromise between the agreement quality for both the bias and opinion labels, assuming that the annotation quality stays the same. To show the difference to 8 annotators, and as an outlook into future extensions of the data set, we also release the codings made by 8 raters\footnote{But recommend to use 5-person ratings when using the full data set.}. We will also add detailed statistics and results about all data and point out our selection process more clearly.  As resources and time were limited, we leave the inclusion of further annotators and more sentences to future work. All raters were master students with a background in Data Science, Computer Science, Psychology, or Intercultural Communication. The groups and their annotators are described in detail in the repository mentioned in Section \ref{Intro}.

\subsection{Evaluation of Data Sets} 

The raw labels obtained during the annotation phase were processed as follows. We calculated an aggregated bias/opinion label for every sentence based on a majority vote principle. For instance, if a sentence was labeled as biased by more than four expert annotators in SG1, we assigned the label \textit{biased} to the sentence. Otherwise, the sentence was marked as \textit{non-biased}.\footnote{Note: In SG2, the threshold reduced respectively due to the lower number of expert annotators.} The annotators did not agree on a label (no majority vote) in some sentences. Here, we assigned the label \textit{no agreement}.

Our annotation scheme allows respondents to mark biased words. In SG1, a word is marked as biased if at least three annotators label it as such. In SG2, the threshold is subsequently reduced to two expert annotators labeling a word as biased.\footnote{We manually inspected all instances to determine reasonable thresholds.} We compute agreement metrics on the sentence level to acquire knowledge about data quality resulting from all annotation approaches. Our agreement metric choice is Krippendorff's $\alpha$ \cite{krippendorff2011computing}, which is a robust agreement metric for studies including varying numbers of annotators per text instance \citep{antoine-etal-2014-weighted}.

We first compared the annotations resulting from MBIC's crowdsourcing approach with our expert-based approach, including eight annotators labeling 1,700 sentences (SG1). 
Table \ref{results_table} shows the %resulting
agreement scores for the bias and opinion labels on a sentence level. Considering the bias agreement, SG1 exhibits fair agreement ($\alpha$ = 0.39) and outperforms MBIC's agreement score ($\alpha$ = 0.21).\footnote{The scoring interpretations are based on guidelines published by \citet{landis1977measurement}.} A similar pattern can be observed regarding the opinion labels (i.e., SG1: $\alpha$ = 0.46; MBIC: $\alpha$ = 0.26). 
Furthermore, MBIC's crowdsourcers labeled more words as biased compared to SG1's experts, i.e., 3,283 vs. 1,530 (absolute) and 2.40 vs. 1.95 (average per biased sentence). Even though media bias detection is generally a difficult task, our inter-annotator agreement is much higher than in existing research in the domain, where $\alpha$ ranges between 0 and 0.20, as shown in Section \ref{sec:relatedwork}.

\begin{table}[t]
\centering
\resizebox{0.4\textwidth}{!}{
\begin{threeparttable}
\begin{tabular}{lcc}
\hline
 \multirow{2}{*}{\textbf{Metric}} & \multicolumn{2}{c}{\textbf{Data}}\\
 \cline{2-3}
 & SG1 & MBIC \\
 \hline
Bias Agreement\textsuperscript{1} & 0.39 & 0.21\\
Opinion Agreement\textsuperscript{1} & 0.46 & 0.26 \\
Total Biased Words & 1530\textsuperscript{3} & 3283\textsuperscript{3} \\
$\varnothing$ Biased Words \textsuperscript{2} & 1.95 & 2.40 \\
\hline
\end{tabular}
\begin{tablenotes}
\small
\item[1] Calculated based on Krippendorff's $\alpha$
\item[2] Average of bias words per biased sentence 
\item[3] Out of 56,826 words in total
\end{tablenotes}
\end{threeparttable}
}
\caption{\label{results_table} Annotation results for the expert-annotated (SG1)  and crowdsourced (MBIC) approach based on 1,700 sentences.}
\end{table}

Table \ref{label_dist} shows the label distribution comparison between SG1 and MBIC.\footnote{Absolute numbers for all labels are reported in the code files at the repository mentioned in Section \ref{Intro}.} We can observe that our expert annotators (SG1) are more conservative in their annotation than the crowdsourcers (MBIC). In the expert data, 43.88\% of the sentences are labeled as biased, whereas the crowdsources annotated 59.88\%. The opinion labels' distribution is fairly balanced in both the expert annotator and crowdsourced data. Factual sentences occur slightly more often than  opinionated sentences in both data sets.

\begin{table}
\centering
\resizebox{0.36\textwidth}{!}{
\begin{threeparttable}
\begin{tabular}{lcc}
\hline
  \multirow{2}{*}{\textbf{Label}} & \multicolumn{2}{c}{\textbf{Data}}\\
 \cline{2-3}
 & SG1 & MBIC \\
 \hline
Biased & 43.88\% & 59.88\%\\
Non-biased & 47.05\% & 31.35\% \\
No agreement & 9.05\% & 8.76\% \\
\cdashline{1-3}
Opinionated & 25.00\% & 30.65\%\\
Factual & 37.59\% & 33.65\% \\
Mixed & 26.64\%  & 25.47\% \\
No agreement & 10.76\%  & 10.24\% \\
\hline
\end{tabular}
\end{threeparttable}
}
\caption{\label{label_dist} Class distribution for SG1's and MBIC's 1700 sentences. }
\end{table}

Next, we evaluate our expert-based annotation approach, including five expert annotators labeling 3,700 sentences (SG2) in comparison to 1,700 (SG1). We compare metrics between both approaches to ascertain whether the reduced number of annotators in SG2 has a substantial impact on the annotator agreement. The finding could yield implications for future research on our extended dataset (SG2).
Table \ref{results_table2} shows agreement metrics for the bias and opinion labels of both expert-annotated approaches, and Table \ref{results_table3} represents label distributions. SG2 exhibits moderate agreement ($\alpha$ = 0.40) in the bias annotation task, and slightly outperforms SG1 ($\alpha$ = 0.39). Regarding the opinion labels, we observe a similar pattern, with SG2 outperforming SG1 more substantially (SG2: $\alpha$ = 0.60; SG2: $\alpha$ = 0.46). The expert annotators of SG1 are more conservative in labeling bias than SG2 (SG1: 43.88\% vs. SG2: 49.26\% labeled as biased).\footnote{Due to the uneven number of annotators in SG2, "no agreement" cases do not exist here.} The opinion labels are distributed marginally skewed in both annotator groups.
Factual sentences occur more often than  opinionated sentences in both data sets.

Further statistics on SG 1 and SG 2 such as bias/opinion distribution per news outlet and topic, the connection between bias and opinion, and the overall topic distribution are provided in the repository mentioned in Section \ref{Intro}.

\begin{table}[t]
\centering
\resizebox{0.4\textwidth}{!}{
\begin{threeparttable}
\begin{tabular}{lcc}
\hline
   \multirow{2}{*}{\textbf{Metric}} & \multicolumn{2}{c}{\textbf{Data}}\\
 \cline{2-3}
 & SG1 & SG2 \\
 \hline
Bias Agreement\textsuperscript{1} & 0.39 & 0.40\\
Opinion Agreement\textsuperscript{1} & 0.46 & 0.60 \\
\hline
\end{tabular}
\begin{tablenotes}
\small
\item[1] Calculated based on Krippendorff's $\alpha$
\end{tablenotes}
\end{threeparttable}
}
\caption{\label{results_table2}Data set annotation results for the expert-based approaches (left: eight annotators labeling 1,700 sentences (SG1); right: five annotators labeling 3,700 sentences (SG2)).}
\end{table}

\begin{table}[t]
\centering
\resizebox{0.4\textwidth}{!}{
\begin{threeparttable}
\begin{tabular}{lcc}
\hline
   \multirow{2}{*}{\textbf{Label}} & \multicolumn{2}{c}{\textbf{Data}}\\
 \cline{2-3}
 & SG1 & SG2 \\
 \hline
Biased & 43.88\%& 49.26\%\\
Non-biased & 47.05\% & 50.70\% \\
No agreement & 9.05\% & 0.00\% \\
\cdashline{1-3}
Opinionated & 25.00\% & 23.35\%\\
Factual & 37.59\% & 43.54\% \\
Mixed & 26.64\%  & 27.21\% \\
No agreement & 10.76\%  & 5.88\% \\
\hline
\end{tabular}
\end{threeparttable}
}
\caption{\label{results_table3}Data set class distribution for the expert-based approaches (left: eight annotators labeling 1,700 sentences (SG1); right: five annotators labeling 3,700 sentences (SG2)).}
\end{table}

%%%%%%%%%%%%%%%%%%%%%%%%%%%%%%%%%%%%%%%%%%%%%%%%%%%%%%%%%%
%%% METHODOLOGY
%%%%%%%%%%%%%%%%%%%%%%%%%%%%%%%%%%%%%%%%%%%%%%%%%%%%%%%%%%

\section{Methodology}\label{sec:methodology}
We propose the use of neural classifiers with automated feature learning capabilities to solve the given media bias classification task. A distant supervision framework, similar to \citet{tang2014}, allows us to pre-train the feature extraction algorithms leading to improved language representations, thus, including information about a sample's bias. As obtaining large amounts of pre-training labeled data using humans is prohibitively expensive, we resort to noisy yet abundantly available labels that provide supervisory signals.

\subsection{Learning Task}
Given a corpus $X$ and a randomly sampled sequence of tokens $x_i \in X$ with $i \in \{1,...,N\}$, the learning task consists of assigning the correct label $y_i$ to $x_i$ where $y_i \in \{0,1\}$ represents the \textit{neutral} and \textit{biased} classes, respectively. The supervised task can be optimized by minimizing the binary cross-entropy loss
\begin{equation}
\label{eq:loss}
\mathcal{L} := - \frac{1}{N} \sum_{i=1}^{N} \sum_{k=\{0,1\}} f_k(x_i) \cdot log(\hat{f_k}(x_i)).
\end{equation}
where $f_k(\cdot)$ is a binary indicator triggering 0 in the case of neutral labels and 1 in the case of a biased sequence. $\hat{f}_k(\cdot)$ is a scalar representing the language model score for the given sequence. 

\subsection{Neural Models}
We fit $\hat{f}_k(\cdot)$ using a range of state-of-the-art language models. Central to the architectural design of these models is \citet{vaswani2017}'s encoder stack of the Transformer relying solely on the attention mechanism.
Specifically, we use the BERT model \citep{devlin2018} and its variants DistilBERT \citep{sanh2019} and RoBERTa \citep{liu2019} that learned bidirectional language representations from the unlabeled text. DistilBERT is a compressed model of the original BERT, and RoBERTa uses a slightly different loss function with more training data than its predecessor. We also evaluate models built on the transformer architecture but differ in the training objective. While DistilBERT and RoBERTa use masked language modeling as a pre-training task, ELECTRA \citep{clark2020} uses a discriminative approach to learn language representations. We also include XLNet \citep{yang2019} in our comparison as an example of an autoregressive model. We systematically evaluate the models' performance on the media bias sentence classification task. We also investigate the impact of an additional pre-training task introduced in the next section on the BERT and RoBERTa models' classification capabilities.

\subsection{Distant Supervision}
Fine-tuning general language models on the target task has proven beneficial for many tasks in NLP \cite{howard2018}. The language model pre-training followed by fine-tuning allows models to incorporate the idiosyncrasies of the target corpus. For text classification, the authors of ULMFiT \cite{howard2018} demonstrated the superiority of task-specific word embeddings. Before fine-tuning, we introduce an additional pre-training task to improve feature learning capabilities considering media bias content. 
The typical unsupervised setting used in the general pre-training stage does not include information on language bias in the learning of the embedded space. To remedy this, we incorporate bias information directly in the loss function (equation \ref{eq:loss}) via distant supervision. In this approach, distant or \textit{weak} labels are predicted from noisy sources, alleviating the need for data labeled by humans. Results by \citet{severyn2015} and \citet{deriu2017} demonstrated that pre-training on larger distant datasets followed by fine-tuning on supervised data yields improved performance for sentiment classification.

A pre-training corpus is compiled consisting of news headlines of outlets with and without a partisan leaning to learn bias-specific word embeddings. The data source, namely, the news outlets, are leveraged to provide distant supervision to our system. As a result, the large amounts of data necessary to learn continuous word representations are gathered by mechanical means alleviating the burden of collecting expensive annotations. The assumption is that the distribution of biased words is denser in some news sources than in others. Text sampled from news outlets with a partisan leaning according to the Media Bias Chart is treated as biased. Text sampled from news organizations with high journalistic standards is treated as neutral. Thus, the mapping of bias and neutral labels to sequences is automatized. The data collection resembles the collection of the ground-truth data described in Section \ref{sec:data}. The defined keywords reflect contentious issues of the US society, as we assume slanted reporting to be more likely among those topics than in the case of less controversial topics. The obtained corpus consisting of 83,143 neutral news headlines and 45,605 biased instances allows for the encoding of a sequence's bias information in the embedded space. The news headlines corpus serves to learn more effective language representations, it is not suitable for evaluation purposes due to its noisy nature. We ensure that no overlap exists between the distant corpus and BABE to guarantee model to guarantee model integrity with respect to training and testing.

%%%%%%%%%%%%%%%%%%%%%%%%%%%%%%%%%%%%%%%%%%%%%%%%%%%%%%%%%%
%%% EXPERIMENTS or Evaluation
%%%%%%%%%%%%%%%%%%%%%%%%%%%%%%%%%%%%%%%%%%%%%%%%%%%%%%%%%%

\section{Experiments}\label{sec:experiments}

\textbf{Training Protocol.} We implement the neural models with HuggingFace's Transformer API \citep{wolf2020}. The model components are instantiated with their pre-trained parameters. Parameters of the classification components are uniformly instantiated and learned. First, we fine-tune and evaluate neural models on BABE. Second, we identify the best performing model of the first run and include the distant supervision pre-training task. 

\textbf{Implementation.} The hyperparameters remain unchanged for pre-training on the distant corpus and fine-tuning on BABE. Sentences are batched together with 64 sentences per mini-batch because estimating gradients in an online learning situation resulted in less stable estimates. 
To optimize $\mathcal{L}$, we use the Adam optimization with a learning rate of $5^{-5}$ \citep{kingma2014}. Training on the distantly labeled corpus is performed for one epoch. While training on BABE, convergence can be observed after three to four epochs. A monitoring system is in place that stops training after two epochs without improvement of the loss and restores the parameters of the best epoch.
All computations were performed on a single Tesla T4 GPU. All in all, pre-training and training of all models is executed in 5 hours.

\textbf{Baseline.} To assess the benefit of modern language models for the domain of media bias, we compare their performance to a traditional feature-based model (Baseline). We use the work by \citet{Spinde2021} as our baseline method, as it offers the most complete set of features for the media bias domain. The authors use syntactic and lexical features related to bias words such as dictionaries of opinion words \citep{hu2004}, hedges \citep{hyland2018a} and assertive and factive verbs \citep{hooper1975a}. \citet{Spinde2021}'s classifier serves as a baseline to evaluate our approach. As feature-based models operate on the word level, we provide comparability by implementing the classification rule that the presence of a predicted biased word leads to the overall sentence being labeled as biased. In contrast, if the baseline model does not label words as biased in a given sequence, the sequence will be classified as neutral.

\textbf{Evaluation Metric.} Given the relatively small size of 3,700 sequences in BABE, we report performance metrics averaged on a 5 fold cross-validation procedure to stabilize the results. Because the class distribution in SG1 is slightly unbalanced, we use stratified cross-validation to preserve this imbalance in each fold. Following the standard in the literature, we report a weighted average of $F_1$-scores.

%%%%%%%%%%%%%%%%%%%%%%%%%%%%%%%%%%%%%%%%%%%%%%%%%%%%%%%%%%
%%% RESULTS
%%%%%%%%%%%%%%%%%%%%%%%%%%%%%%%%%%%%%%%%%%%%%%%%%%%%%%%%%%

\section{Results}\label{sec:results}
Table \ref{Tab:class_res} summarizes our performance results. Our baseline using engineered features exhibits low scores of 0.511 and 0.569 for SG1 and SG2, respectively.\footnote{In this Section, we show three decimal places to account for detailed model differences.} BERT improves over the baseline by a large margin of 0.251 points on SG1 and 0.220 points on SG2. DistilBERT exhibits a lower performance for both corpora, whereas RoBERTa is the strongest representative of BERT-based models. Both models based on a different training approach than BERT, namely ELECTRA and XLNet, do not match the performance of BERT and its optimized variants. These results reaffirm established findings of the attention mechanism's advantage over traditional models \cite{e23030283} and indicate the benefits of large pre-trained models' for media bias detection. 

Models trained and evaluated on SG2 generally perform better due to their bigger corpus size. The increase is around 0.02 points of the macro $F_1$-score for all models except RoBERTa + distant, where it is insignificant. Overall, we believe the improvement indicates that extending the data set in the future will be valuable.

Results of the fourth block of table \ref{Tab:class_res} show that the distant supervision pre-training task leads to an improvement over BERT and RoBERTa. Our best performing model BERT + distant on SG2 achieves a macro $F_1$-score of 0.804 and improves over the BERT model by 0.02 points. Media bias can be better captured when word embedding algorithms are pre-trained on the news headlines corpus with distant supervision based on varying news outlets. With the added data, information on a sequence's bias is incorporated in the loss function, which is not the case in "general purpose" language models.

\begin{table}[ht]
\centering
\resizebox{0.49\textwidth}{!}{
\begin{threeparttable}
\begin{tabular}{p{3cm}cc}
\hline
\multirow{2}{*}{\textbf{Model}}  & \multicolumn{2}{c}{\textbf{Macro $F_1$}}  \\
 \cline{2-3} 
 & SG1 & SG2 \\
\hline
Baseline & 0.511 (0.008) & 0.569 (0.008)  \\
\cdashline{1-3}
BERT & 0.762 (0.019) & 0.789 (0.011) \\
DistilBERT & 0.758 (0.029) & 0.777 (0.009)\\
RoBERTa & 0.775 (0.023) & 0.799 (0.011)\\
\cdashline{1-3}
ELECTRA & 0.742 (0.020) & 0.760 (0.013)\\
XLNet & 0.760 (0.042) & 0.797(0.015) \\
\cdashline{1-3}
BERT + distant & 0.778 (0.017) & \textbf{0.804} (0.014) \\
RoBERTa + distant & \textbf{0.798} (0.022) & 0.799 (0.017)\\
\hline
\end{tabular}
\begin{tablenotes}
\small
\item Standard errors across folds in parentheses. 
\item The first model block shows the best results of feature-based models. The second block of models consists of BERT and optimize variants. The models in the third block use new architectural or training approaches. The fourth block refers to models having learned bias-specific embeddings from the distantly supervised corpora.
\item The best results are printed in \textbf{bold}.
\end{tablenotes}
\end{threeparttable}
}
\caption{Stratified 5 fold cross-validation results.}
\label{Tab:class_res}
\end{table}

%%%%%%%%%%%%%%%%%%%%%%%%%%%%%%%%%%%%%%%%%%%%%%%%%%%%%%%%%%
%%% FINAL CONSIDERATIONS
%%%%%%%%%%%%%%%%%%%%%%%%%%%%%%%%%%%%%%%%%%%%%%%%%%%%%%%%%%

\section{Discussion}\label{sec:discussion}
Employing annotators with domain expertise allows us to achieve an inter-annotator agreement of $\alpha$ = 0.40, which is higher than existing data sets \cite{spinde2021mbic}. We believe domain knowledge and training alleviate the difficulty of identifying bias and are imperative to create a strong benchmark due to the complexity of the task. In future work, apart from improving the current data set and classifier, we will also explore why a text passage might be biased, not just its overall classification. Currently, traditional machine learning models are interpretable \cite{Spinde2021} but outperformed by recurrence and attention-based models. Hand-crafted features like static dictionaries cannot adequately address the complexity and context-dependence of bias. 

We argue that standard metrics (e.g., accuracy and $F_1$) provide a limited perspective into a model's predictive power in case of a complex construct like media bias. Further research needs to tackle these pitfalls to propose systems with better generalization capabilities. A promising starting point might be a more refined evaluation scheme that decomposes the bias detection task into multiple sub-tasks, such as presented in CheckList \citep{ribeiro2020beyond}. This scheme will also allow us to understand how our system performs on different types of bias (e.g., bias by context, by linguistics, by overall reporting). Additionally, we believe that current research on explainable artificial intelligence might increase users' trust in neural-based classifiers. Existing research already presents ways to visualize Transformer-based models and make their results more accessible and interpretable \citep{vig2019}. Lastly, combining neural methods with advances in linguistic bias theory \cite{Spinde2021} to explain a classifier's decision to users will also be part of our future work. 

For this work, we focused on sentence level bias, which is often used in the media bias domain. Still, in addition to the 3,700 labeled sentences, we also include word level annotations in our data set to encourage solutions focusing on more granular characteristics. We believe that word level bias conveys strong explanatory and structural knowledge and see a detailed word level bias analysis and detection as a promising research direction.

\section{Conclusion}\label{sec:conclusion}
This work proposes BABE, a new high-quality media bias data set. BABE contains 3,700 labeled sentences, and enables us to compare crowdsourcing and expert annotations directly. Additionally, we propose a sentence level bias classifier based on BERT, which outperforms existing work in the domain. By deriving bias-specific word embeddings using distant supervision, we have improved our classifier even more, achieving a macro $F_1$-score = 0.804. We make all models, data, and code publicly available.\footnote{We publish the link in Section \ref{Intro}.}

%%%%%%%%%%%%%%%%%%%%%%%%%%%%%%%%%%%%%%%%%%%%%%%%%%%%%%%%%%
%%% ETHICAL CONSIDERATIONS
%%%%%%%%%%%%%%%%%%%%%%%%%%%%%%%%%%%%%%%%%%%%%%%%%%%%%%%%%%
\section*{Ethics/Broader Impact Statement}
Detecting and highlighting media bias instances may have many positive implications and can mitigate the effects of such biases \cite{baumer2015a}. Still, bias is a highly sensitive topic, and some forms of bias especially rely on other factors than the content itself, such as a different perception of any text related to the individual background of a reader. When showing detected bias or news outlet classifications on a political or polarization scale to a reader, every algorithm should be transparent in how the classifications were made. In general, the topic should be handled carefully. We want to point out that it is uncertain if and how actual news consumers would like to obtain such information. Some research groups working on the detection of bias have also started to work on psychological and societal questions related to bias \cite{spinde2020b}. From a social science perspective, it remains to be explored how a classifier can mitigate the negative effects of biased media on society.

Generally, when performed in a balanced and transparent way, bias detection might positively affect collective decision-making and opinion formation processes. As such, and to this point, we see no immediate negative ethical or societal impacts of our work beyond what applies to other core building blocks of deep learning. Apart from the system transparency, as mentioned above, one important factor to consider when building, training, and presenting any media bias classifier is a manipulation protection strategy. Participants in any study, especially public ones, should not be able to tweak algorithms and therefore, e.g., flag neutral content as biased to undermine the validity of media bias detection systems. Hence, annotations should always be compared among multiple users, where trustworthiness can at least be largely assured. In open (crowdsourcing) scenarios, collecting user characteristics and consciously implementing specific content (like questions that should give an obvious answer but might be answered differently when users a following any pattern) is important.

As a side effect of our project, we experienced that our annotators learned to read the news more critically and reflected more about what they read even after the study ended. We have already started to implement the insights we gained into ways to improve the perception of bias in a game, teaching players to read news with greater care and execute a large study investigating how such a game can affect children, especially in school.

Our data set is completely anonymized to preserve the identities of everyone involved.

%%%%%%%%%%%%%%%%%%%%%%%%%%%%%%%%%%%%%%%%%%%%%%%%%%%%%%%%%%
%%% ACKNOWLEDGMENTS
%%%%%%%%%%%%%%%%%%%%%%%%%%%%%%%%%%%%%%%%%%%%%%%%%%%%%%%%%%

\section*{Acknowledgments}

The Hanns-Seidel-Foundation, Germany, supported this work, as did the DAAD (German Academic Exchange Service). We are grateful to our raters, who we will keep anonymous, as we are grateful to Dr. Franz Hahn and Prof. Dr. Jelena Mitrović for helping us to hire a part of the annotators.

%\printbibliography[keyword=primary]
\bibliography{anthology,short}
\bibliographystyle{acl_natbib}

%\appendix

%\section{Example Appendix}
%\label{sec:appendix}

%This is an appendix.

\end{document}